\documentclass[letterpaper]{article} 
\usepackage{aaai23}  
\usepackage{times}  
\usepackage{helvet}  
\usepackage{courier}  
\usepackage{amssymb}
\usepackage{makecell}
\usepackage{enumerate}
\usepackage{amsmath}
\usepackage{booktabs}
\usepackage[hyphens]{url}  
\usepackage{graphicx} 
\usepackage{natbib}  
\usepackage{caption} 
\usepackage{algorithm}
\usepackage{algorithmic}
\usepackage{graphicx} 

\usepackage{float} 
\usepackage{listings}
\DeclareCaptionStyle{ruled}{labelfont=normalfont,labelsep=colon,strut=off} 
\floatstyle{ruled}
\newfloat{listing}{tb}{lst}{}
\floatname{listing}{Listing}

\newtheorem{corollary}{Corollary}
\newtheorem{thm}{Theorem}
\title{Combining Adversaries with Anti-adversaries in Training}
\author{
    Xiaoling Zhou,
    Nan Yang,
    Ou Wu \thanks{Corresponding author.}
}
\affiliations{
    Center for Applied Mathematics, Tianjin University, China\\

    \{xiaolingzhou, yny, wuou\}@tju.edu.cn
}

\usepackage{bibentry}

\begin{document}

\nocopyright

\renewcommand\thethm{\arabic{thm}}
\setcounter{thm}{0}
\maketitle
\begin{abstract}
Adversarial training is an effective learning technique to improve the robustness of deep neural networks. In this study, the influence of adversarial training on deep learning models in terms of fairness, robustness, and generalization is theoretically investigated under more general perturbation scope that different samples can have different perturbation directions (the adversarial and anti-adversarial directions) and varied perturbation bounds. Our theoretical explorations suggest that the combination of adversaries and anti-adversaries (samples with anti-adversarial perturbations) in training can be more effective in achieving better fairness between classes and a better tradeoff between robustness and generalization in some typical learning scenarios (e.g., noisy label learning and imbalance learning) compared with standard adversarial training. On the basis of our theoretical findings, a more general learning objective that combines adversaries and anti-adversaries with varied bounds on each training sample is presented. Meta learning is utilized to optimize the combination weights. Experiments on benchmark datasets under different learning scenarios verify our theoretical findings and the effectiveness of the proposed methodology.
\end{abstract}


\section{Introduction}
\noindent Apart from the standard generalization error (also known as natural error), robust generalization error (also known as robust error) has received great attention in recent years. A deep neural network with a low robust error can cope well with adversarial attacks. Adversarial training is an effective technique to reduce the robust error of a model~\cite{wong2020fast,review}. Given a model $f(\cdot)$ and a sample $\boldsymbol{x}$ associated with a label $y$, classical adversarial training methods~\cite{pgd,FGSM} first generate an adversary (i.e., adversarial example) $\boldsymbol{x}_{\text{adv}}$ for $\boldsymbol{x}$ with the following optimization:
\begin{equation}
{\boldsymbol{x}_{\text{adv}}} =\boldsymbol{x} + \arg \mathop {\max }\limits_{\left\| \boldsymbol{\delta}  \right\| \le \epsilon } \ell(f(\boldsymbol{x} + \boldsymbol{\delta} ),y),
\label{adv}
\end{equation}
where $\ell(\cdot,\cdot)$ is a loss function, $\boldsymbol{\delta}$ is the perturbation term, and $\epsilon$ is the perturbation bound. Adversaries are then leveraged as the training data to learn a more robust model. A number of variations for adversarial training have been proposed in recent literature. Zhang et al.~\shortcite{zhangtrades} decomposed the robust error into the natural and boundary errors. They developed a new method, namely, TRADES, to obtain a better tradeoff between standard generalization and robustness. Wang et al.~\shortcite{wang2020MADV} proposed a misclassification-aware adversarial training method to focus on the misclassified examples.

In addition to the design of new methods, theoretical studies have been conducted to explore the effectiveness and ineffectiveness of adversarial training~\cite{review}. Yang et al.~\shortcite{closerlook} concluded that existing adversarial methods cannot achieve an ideal tradeoff between accuracy and robustness due to the insufficient smoothness~\cite{smooth} and generalization
properties of classifiers trained by these methods. They pointed out that customized optimization methods or better network
architectures should be proposed. Xu et al.~\shortcite{xu2021robust} revealed that adversarial training introduces severe unfairness between different categories. Thus, they developed a new method that sets varied perturbation bounds for each class, resulting in better fairness. Different from these studies, we conjectured that one possible reason leading to unsatisfied tradeoff and fairness is that not all training samples should be perturbed adversarially. For instance, adversaries of noisy samples may harm the model performance~\cite{noise_adv}, and these samples should be perturbed in the anti-adversarial direction to reduce their negative influence on model optimization. Zhu et al.~\shortcite{interation} re-annotated pseudo labels for possible noisy samples before generating adversaries for them. The generated adversaries are actually perturbed anti-adversarially in binary classification tasks.
In this study, samples with anti-adversarial perturbations are called anti-adversaries\footnote{The anti-adversary defined by Alfarra et al.~\shortcite{combating} is different from ours. They utilize anti-adversaries to deal with attacks,
whereas we aim to improve robustness, accuracy, and fairness.} ($\boldsymbol{x}_\text{at-adv}$) 
\begin{equation}
{\boldsymbol{x}_{\text{at\text{-}adv}}} =\boldsymbol{x} + \arg \mathop {\min }\limits_{\left\| \boldsymbol{\delta}  \right\| \le \epsilon } \ell(f(\boldsymbol{x} + \boldsymbol{\delta}),y).
\label{at-adv}
\end{equation}

This study conducts a comprehensive theoretical analysis of adversarial training in the presence of two different perturbation directions (adversarial and anti-adversarial) and varied bounds. Several typical learning scenarios are considered, including classes with different learning difficulties, imbalance learning, and noisy label learning. Our theoretical findings reveal that the perturbation directions and bounds can remarkably influence the model training. The combination of adversaries and anti-adversaries with varied bounds can improve the fairness among classes and achieve a better tradeoff between accuracy and robustness. Accordingly, a general objective that combines adversaries and anti-adversaries is constructed for adversarial training. A meta learning-based method is then proposed to optimize this objective, in which the perturbation direction and bound of each training sample is adjusted in accordance with its learning characteristics during training. Our experimental results show that the combining strategy outperforms state-of-the-art adversarial training methods. Our experimental observations are in accordance with our theoretical findings.  

The contributions of our study are as follows:
\begin{itemize}
\item To the best of our knowledge, this is the first work that combines adversaries and anti-adversaries in training. A comprehensive theoretical analysis is conducted for the role of the combination strategy with varied perturbation bounds\footnote{Existing theoretical studies presume that the perturbation bounds are identical for all training samples.} under three typical learning scenarios. 
\item A new objective is established for adversarial training by combining adversaries and anti-adversaries.
Meta learning is utilized to solve the optimization, and the perturbation direction and bound for each training sample are determined in accordance with its learning characteristics, such as training loss and margin.
\end{itemize}
\begin{figure*}[t] 
\centering
\includegraphics[width=1\textwidth]{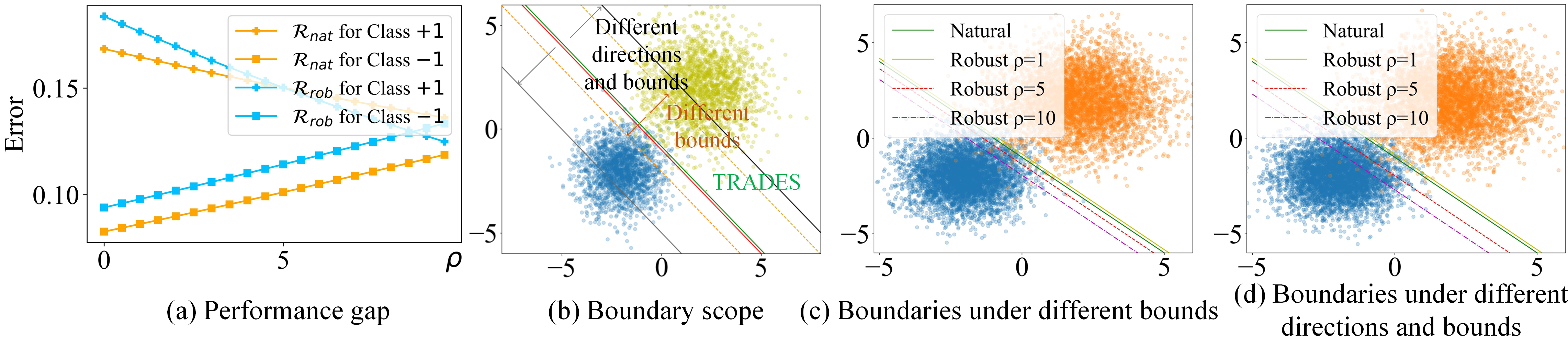}
\caption{(a) Variation of performance gaps between classes as $\rho$ increases. (b) Scope of the classification boundary of different manners. The values of parameters are $K\!=\!2$, $\eta\!=\!2$, $\epsilon\!=\!0.2$, and $\sigma\!=\!1$. The bounds for class ``$+1$" and ``$-1$" are denoted as $\rho_{+}\times\epsilon$ and $\rho_{-}\times\epsilon$ ($-\eta/\epsilon<\rho_+, \rho_-<\eta/\epsilon$), respectively. $\rho_+(\rho_-)<0$ denotes that class ``$+1(-1)$" is anti-adversarially perturbed. The online material provides the formulas of boundaries. (c) Logistic regression classifier boundaries (natural and robust) on simulated data in Eq.~(\ref{data_distri}). (d) Logistic regression classifier boundaries (natural and robust with different directions).}
\end{figure*}
\section{Related Work}
\subsection{Tradeoff and Fairness in Adversarial Training}
Recent studies on adversarial training focus on the tradeoff between accuracy and robustness.
Efforts~\cite{adv_hurt_genera,zhangtrades,tradeoff1,Yang} have been made to reduce the natural errors of the adversarially trained models, such as adversarial training with semi/unsupervised learning and robust local feature~\cite{Songet}.
Rice et al.~\shortcite{Riceet} systematically investigated the role of various techniques used in deep learning for achieving a better tradeoff, such as cutout, mixup, and early stopping, where early stopping is found to be the most effective. This investigation was also confirmed by Pang et al.~\shortcite{Tianyu}.
Unfairness is also a problem caused by adversarial training.
Xu et al.~\shortcite{xu2021robust} trained a robust classifier to minimize error and stressed it to satisfy two fairness
constraints. Several studies~\cite{Ding2020MMA,Cheng2020CAT,Balaji2019Instance} adaptively tune the perturbation bounds for each sample with the inspiration that samples near the decision boundary should have small bounds.

\subsection{Meta Learning}
Meta learning has aroused great interest in recent years. Existing meta learning methods can be divided into three categories, namely, metric-based~\cite{metametric1,metametric2}, model-based~\cite{metamodel1}, and optimizing-based~\cite{MAML,metaoptimization2} methods. The algorithm we adopted that is inspired by Model-Agnostic Meta-Learning~\cite{MAML} belongs to the optimizing-based methods. The data-driven manner of meta optimization is always utilized to learn the sample weights or the hyperparameters~\cite{L2RW,shu2019meta}.



\section{Theoretical Investigation}
This section conducts theoretical analyses to assess the influence of two different perturbation directions and varied bounds on adversarial training in three typical binary classification cases. Proofs are presented in the online material.
\subsection{Notation}
We denote the sample instance as $\boldsymbol{x}\in\mathcal{X}$ and $y\in\mathcal{Y}$ as the label, where $\mathcal{X}\subseteq\mathbb{R}^{d}$ indicates the instance space, and $\mathcal{Y}=\{-1,+1\}$ indicates the label space. The classification model $f$ is a mapping from the input data space $\mathcal{X}$ to the label space $\mathcal{Y}$.
It can be parametrized by using linear classifiers or deep neural networks. The overall natural error of $f$ is denoted as $\mathcal{R}_{\text{nat}}(f)\!:=\! \operatorname{Pr}(f(\boldsymbol{x})\!\neq\!y)$. 
The overall robust error is denoted as $\mathcal{R}_{\text {rob}}(f)\!:=\!\operatorname{Pr}(\exists \boldsymbol{\delta}~ ||\boldsymbol{\delta}||\!\le\!\epsilon, \text{s.t.} f\left(\boldsymbol{x}+\boldsymbol{\delta}\right)\!\neq\! y)$. 

\subsection{Case I: Classes with Different Difficulties}
In this case, the binary setting established by Xu et al.~\shortcite{xu2021robust} is followed. 
The data from each class follow a Gaussian distribution $\mathcal{D}$ that is centered on $\boldsymbol{\theta}$ and  $\boldsymbol{-\theta}$, respectively. A $K$-factor difference is found between two classes’ variances: $\sigma_{+1}: \sigma_{-1}=K: 1$ and $K>1$. The data follow
\begin{equation} 
\begin{aligned}
y \stackrel{u . a . r}{\sim}\{-1,+1\}, \quad \boldsymbol{\theta}=({\eta, \ldots, \eta} ) \in \mathbb{R}^d,\eta > 0,\\
\boldsymbol{x} \sim\left\{\begin{array}{ll}\mathcal{N}\left(\boldsymbol{\theta}, \sigma_{+1}^{2} \boldsymbol{I}\right), & \text { if } y=+1, \\ \mathcal{N}\left(-\boldsymbol{\theta}, \sigma_{-1}^{2} \boldsymbol{I}\right), & \text { if } y=-1.\end{array}\right.
\end{aligned}
\label{data_distri}
\end{equation}
Class ``$+1$" is harder because the optimal linear classifier will give a larger error to class ``$+1$" than class ``$-1$". Xu et al.~\shortcite{xu2021robust} proved that adversarial training with an equal bound will exacerbate the performance gap (including natural and robust errors) between classes and hurt the harder class. We show that adversarial training with unequal bounds on two classes can tune the performance gap and the tradeoff between the robustness and accuracy of the model. Let $\sigma_{-1} = \sigma$. The following theorem is first proposed.

\begin{thm}
For a data distribution $\mathcal{D}$ in Eq.~(\ref{data_distri}), assume that the perturbation bounds of class  ``$-1$" and ``$+1$" are $\epsilon$ and $\rho\times\epsilon$ ($0\leq\epsilon,\rho{\epsilon}<{\eta}$), respectively. The natural errors of the optimal robust linear classifier $f_{\text{rob}}$ 
for two classes are
\begin{equation}
\small
\resizebox{1\hsize}{!}{$\begin{aligned} & \mathcal{R}_{\text{nat}}\left(f_{\text{rob}},-1\right) = \operatorname{Pr}\left\{\mathcal{N}(0,1) \leq B-K \cdot \sqrt{B^{2}+q(K)}-\frac{\sqrt{d}}{\sigma} \epsilon\right\}, \\ & \mathcal{R}_{\text{nat}}\left(f_{\text{rob}},+1\right) = \operatorname{Pr}\left\{\mathcal{N}(0,1) \leq-K \cdot B+\sqrt{B^{2}+q(K)}-\frac{\sqrt{d}\rho}{K\sigma}\epsilon\right.\}, \end{aligned}$}
\end{equation}
where $B=\frac{2}{K^{2}-1}\frac{\sqrt{d}(\eta-\frac{\epsilon(1+\rho)}{2})}{\sigma}$, and $q(K)=\frac{2\log K}{K^{2}-1}$.
\end{thm}

The robust errors are shown in the online material. The natural and robust errors change with different $\rho$ values. A corollary is derived in accordance with Theorem~1.

\begin{corollary}
The data and perturbations in Theorem~1 are followed. When $K<\exp({{d(\eta-\epsilon)^2}/{2\sigma^2}})$, the adversarially trained model will increase and decrease the natural and
robust errors of class ``$-1$" and class ``$+1$", with the increase in $\rho$, respectively.
\end{corollary}

Accordingly, the performance gaps of $\mathcal{R}_{\text{nat}}$ and $\mathcal{R}_{\text{rob}}$ decrease with the increase in $\rho$, and better fairness can be achieved, as shown in Fig.~1 (a). In Fig.~1 (c), the boundary shifts toward the easy class ``$-1$". From Fig.~1 (b), adversarial training with varied bounds contributes to larger scope of the boundary compared with TRADES~\cite{zhangtrades}. Thus, a better tradeoff can be attained. Therefore, fairness and tradeoff can be tuned with different $\rho$ values. Next, anti-adversaries are considered. Assume that samples in class ``$-1$" perform anti-adversarial perturbation. Similar to Theorem~1, a theorem calculating the natural and robust errors
is proposed as shown in the online material. A corollary is then derived, indicating that the combination of adversaries and
anti-adversaries can tune the performance gap and tradeoff.
\begin{corollary}
 For a data distribution $\mathcal{D}$ in Eq.~(\ref{data_distri}), assume that class ``$-1$" is anti-adversarially perturbed with the bound $\epsilon$, and class ``$+1$" is adversarially perturbed with the bound $\rho\times\epsilon$ ($0\leq\epsilon,\rho{\epsilon}<{\eta}$). When $K<\exp({{d(\eta-\epsilon)^2}/{2\sigma^2}})$, the adversarially trained model will increase and decrease the natural and robust errors of class ``$-1$" and class ``$+1$", with the increase in $\rho$, respectively.
\end{corollary}

In accordance with Corollaries 1 and 2, the adversarial training and the combination strategy can nearly attain the same performance. However, the combination strategy can contribute to the
largest scope of the boundary, as shown
in Fig.~1 (b). Thus, the combination strategy is more effective in achieving a better tradeoff and fairness theoretically. As shown in Figs.~1 (c) and (d), the combination strategy has a more pronounced effect under the same bound (i.e., the same $\rho$), indicating that it needs smaller bounds when the same performance is achieved. Thus, the combination strategy is more efficient than only the adversarial perturbation, indicating that anti-adversaries are valuable.

\subsection{Case II: Classes with Imbalanced Proportions}
In this case, 
the two variances in Eq.~(\ref{data_distri}) are assumed to be identical\footnote{The case with different variances can be explored similarly.}, that is, $\sigma_{+1} = \sigma_{-1}=\sigma$. However, $p(y=+1)$ ($p_+$) is no longer equal to $p(y=-1)$ ($p_-$). Without loss of generality, let $p_+:p_-=1:V$ and $V>1$.

Class ``$-1$" is the majority category, and an optimal linear classifier will give a smaller natural error for class ``$-1$" than class ``$+1$", as proved in the online material. Similarly, we proved that standard adversarial training will exacerbate the performance gap between classes and hurt the smaller class. We then show that adversarial training with unequal bounds on the two classes will tune the performance gap between classes and the tradeoff between robustness and accuracy. The following theorem is first proposed. 

\begin{thm}
For a data distribution $\mathcal{D}_{V}$ described above with the imbalance factor $V$, assume that the perturbation bounds of classes  ``$-1$" and ``$+1$" are $\epsilon$ and $\rho\times\epsilon$ ($0\leq\epsilon, \rho{\epsilon}<{\eta}$), respectively. The natural errors of the optimal robust linear classifier $f_{\text{rob}}$ 
for the two classes are
\begin{equation}
\resizebox{0.9\hsize}{!}{$\begin{aligned} & \mathcal{R}_{\text{nat}}\left(f_{\text{rob}},-1\right) = \operatorname{Pr}\left\{\mathcal{N}(0,1) \leq -A-\frac{\log V}{2A}-\frac{\sqrt{d}}{\sigma} \epsilon\right\}, \\ & \mathcal{R}_{\text{nat}}\left(f_{\text{rob}},+1\right) = \operatorname{Pr}\left\{\mathcal{N}(0,1) \leq-A+\frac{\log V}{2A}-\frac{\sqrt{d}\rho}{\sigma} \epsilon\right\}, \end{aligned}$}
\end{equation}
where $A={\sqrt{d}(\eta-{\epsilon(1+\rho)}/{2})}/{\sigma}$.
\end{thm}

A corollary is derived on the basis of Theorem~2.
\begin{corollary}
The data and perturbations in Theorem~2 are followed. When $V<\exp({{d(\eta-\epsilon)^2}/{2\sigma^2}})$, the adversarially trained model will increase and decrease the natural and
robust errors of class ``$-1$" and class ``$+1$", with the increase in $\rho$, respectively.
\end{corollary}

From Corollary~3, the performance gaps between classes can be decreased with different $\rho$ values. The boundary can be moved within the scope with different $\rho$ values that covers the boundary of standard adversarial training.
Therefore, a better tradeoff can be attained by adversarial training with varied bounds. Next, the anti-adversaries are considered. We assume that samples in class ``$-1$" perform anti-adversarial perturbation. Similar to Theorem~2, a theorem that portrays the training occasion where the adversaries and anti-adversaries are combined is proposed, as shown in the online material. A corollary is then derived. 

\begin{corollary}
For a data distribution $\mathcal{D}_{V}$ in Theorem~2, the perturbations in Corollary~2 are followed. 
When $V<\exp({{d(\eta-\epsilon)^2}/{2\sigma^2}})$, the adversarially
trained model will increase and decrease the natural and
robust errors of class ``$-1$" and class ``$+1$", with the increase in $\rho$, respectively.
\end{corollary}

In accordance with Corollary~4, the performance gaps between classes can be tuned by the combination strategy. In addition, it can contribute to the larger scope of the classification boundary compared with only adversaries, and a better tradeoff can be attained. When the same performance is achieved, combining adversaries and anti-adversaries has a smaller bound. Therefore, the combination strategy is more efficient than only the adversarial perturbation.
More details are presented in the online material.  

\subsection{Case III: Classes with Noisy Labels}
In this case, 
the two classes’ variances and prior probabilities are assumed to be identical, that is, $\sigma_{+1} = \sigma_{-1}$ and $p_+ = p_-$. Without loss of generality, class ``$-1$" is assumed to contain flipped noisy labels. Two main conclusions are obtained. 1) The adversaries of noisy samples will harm the tradeoff and fairness of the robust model.
2) If noisy samples are anti-adversarially perturbed with a bound $\rho\times\epsilon$ and clean samples are adversarially perturbed with a bound $\epsilon$, then the natural and robust errors of class ``$-1$" and class ``$+1$" will be decreased and increased with the increase in $\rho$, respectively. 
Thus, the combination strategy with varied bounds is effective in achieving a lower performance gap between classes and a better tradeoff between the accuracy and robustness on noisy data. The relevant theorems are shown in the online material.


\subsection{Summarization}
Our theoretical analysis comprehensively reveals that the perturbation directions and bounds remarkably influence the generalization, robustness, and fairness of the robust model under three typical learning scenarios. Adversarial training with different perturbation directions and bounds can better tune the performance gap between classes and the tradeoff between robustness and accuracy. Existing studies ignored anti-adversaries that are valuable. Thus, a new optimized objective considering anti-adversaries is proposed.

\section{Methodology}

Illuminated by the theoretical analysis, a new objective function is first established. Accordingly, a meta learning-based method that \underline{c}ombines \underline{a}dversaries and \underline{a}n\underline{t}i-adversaries (CAAT) in training with a varied bound for each sample is proposed to solve the optimization, as shown in Fig.~2.

\begin{figure}[t] 
\centering
\includegraphics[width=0.47\textwidth]{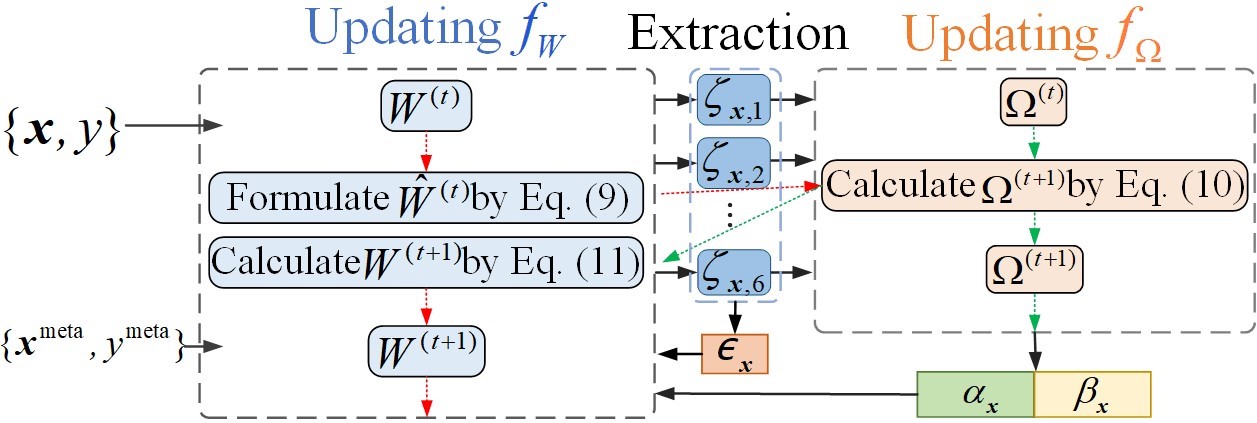}
\caption{Overall structure of CAAT. The red and green lines represent the learning loops of the classifier network and weighting network, respectively.}
\end{figure}
\subsection{Proposed Objective Function}
 
Ideally, the objective function that combines adversaries and anti-adversaries can be formulated as
\begin{equation}
\begin{aligned} 
\min _{{\boldsymbol{W}},\alpha,\beta} &\mathbb{E}_{\boldsymbol{x}}\{\alpha_{\boldsymbol{x}} \ell\!\left(f_{_{\boldsymbol{W}}}\left(\boldsymbol{x}_{\text{adv}}\right),y\!\right)\!+\!\beta_{\boldsymbol{x}} \ell\!\left(f_{_{\boldsymbol{W}}}(\boldsymbol{x}_{\text{at-adv}}),y\!\right)\},\\&
\text{s.t.}~\alpha_{\boldsymbol{x}}+\beta_{\boldsymbol{x}} = 1~\text{and}~ \alpha_{\boldsymbol{x}},\beta_{\boldsymbol{x}}\in\{0,1\},
\end{aligned}
\label{objective}
\vspace{-0.0in}
\end{equation}
where $\boldsymbol{x}_{\text{adv}}$ and $\boldsymbol{x}_{\text{at-adv}}$ are calculated by using Eqs.~(1) and (2) with varied bound $\epsilon_{\boldsymbol{x}}$ for each sample $\boldsymbol{x}$, respectively; $\alpha_{\boldsymbol{x}}$ and $\beta_{\boldsymbol{x}}$ are the combination weights; $f_{_{\boldsymbol{W}}}$ is the classifier network with the parameter $\boldsymbol{W}$. When $\alpha_{\boldsymbol{x}} \equiv 1$, Eq.~(\ref{objective}) can be reduced to the objective of standard adversarial training.

To solve Eq.~(\ref{objective}), we first assume that the values of $\alpha_{\boldsymbol{x}}$ and $\beta_{\boldsymbol{x}}$ depend on the training characteristics of sample $\boldsymbol{x}$. Accordingly, their values are produced by a weighting network $f_{\boldsymbol{\Omega}}$ (parameterized by $\boldsymbol{\Omega}$), where its input is a series of training characteristics $\boldsymbol{\zeta}_{\boldsymbol{x}}$ of $\boldsymbol{x}$ shown in Fig.~2. $\ell(f_{_{\boldsymbol{W}}}(\boldsymbol{x}_\text{adv}),y)$ can be divided into $\ell(f_{_{\boldsymbol{W}}}(\boldsymbol{x}),y)$ and $\ell(f_{_{\boldsymbol{W}}}(\boldsymbol{x}),f_{_{\boldsymbol{W}}}(\boldsymbol{x}_\text{adv}))$ to achieve a better tradeoff between the accuracy and robustness~\cite{zhangtrades}. To improve the fairness among classes, we further stress $f$ to satisfy two fairness constraints following Ref.~Xu et al.\shortcite{xu2021robust}. Thus, our adopted objective function is
\begin{equation}
\begin{aligned}
\min _{{\boldsymbol{W}},{\boldsymbol{\Omega}}} &\mathbb{E}_{\boldsymbol{x}}\{{\alpha_{\boldsymbol{x}}[\ell(f_{_{\boldsymbol{W}}}(\boldsymbol{x}),y)}+{\lambda \ell\left(f_{_{\boldsymbol{W}}}(\boldsymbol{x}),f_{_{\boldsymbol{W}}}\left(\boldsymbol{x}_{\text{adv}}\right)\right)}]\\&+{\beta_{\boldsymbol{x}} \ell\left(f_{_{\boldsymbol{W}}}(\boldsymbol{x}_{\text{at-adv}}),y\right)}\},\\&
\text { s.t. } \left\{\begin{array}{l}
[\alpha_{\boldsymbol{x}}, \beta_{\boldsymbol{x}}] = f_{\boldsymbol{\Omega}}(\boldsymbol{\zeta_{\boldsymbol{x}}}), \forall \boldsymbol{x} \in \mathcal{X}, \\
\mathcal{R}_{\text {nat}}(f_{_{\boldsymbol{W}}}, c)-\mathcal{R}_{\text {nat}}(f_{_{\boldsymbol{W}}}) \leq \tau_{1}, \forall c \in \mathcal{Y},\\
\mathcal{R}_{\text {bdy}}(f_{_{\boldsymbol{W}}}, c)-\mathcal{R}_{\text {bdy}}(f_{_{\boldsymbol{W}}}) \leq \tau_{2},\forall c \in \mathcal{Y},
\end{array}\right.
\end{aligned}
\end{equation}
where $\mathcal{R}_\text{bdy}$ is the boundary error of the model, denoted as $\mathcal{R}_{\text {bdy}}(f_{_{\boldsymbol{W}}})=\operatorname{Pr}(\exists {\boldsymbol{x}_{\text{adv}} \in \mathbb{B}(\boldsymbol{x}, \epsilon)}, f_{_{\boldsymbol{W}}}(\boldsymbol{x}_\text{adv})\neq f_{_{\boldsymbol{W}}}(\boldsymbol{x})\}$; $\mathcal{R}_{\text {nat}}(f_{_{\boldsymbol{W}}}, c)=\operatorname{Pr}(f_{_{\boldsymbol{W}}}(\boldsymbol{x})\neq y\mid y=c\}$; $\mathcal{R}_{\text {bdy}}(f_{_{\boldsymbol{W}}}, c)=\operatorname{Pr}(\exists {\boldsymbol{x}_{\text{adv}} \in \mathbb{B}(\boldsymbol{x}, \epsilon)}, f_{_{\boldsymbol{W}}}(\boldsymbol{x}_\text{adv})\neq f_{_{\boldsymbol{W}}}(\boldsymbol{x})\mid y=c\}$; $f_{\boldsymbol{\Omega}}$ is a multilayer perception (MLP) network with a hidden layer and a $\tau$-softmax layer: $\text{Softmax}((\boldsymbol{h}\boldsymbol{\omega}+\boldsymbol{b})/\tau)$; $\lambda>0$ is a regularization parameter that adjusts the influence of the natural and boundary errors on the model; $\tau_{1}$ and $\tau_{2}$ are small and positive predefined parameters. The approach for solving the two fairness constraints is the same as that in Ref.~Xu et al.\shortcite{xu2021robust}, where a Lagrangian is formed.
\begin{algorithm}[t]
\small
    \caption{CAAT}
    \label{alg1}
    \textbf{Input}: \#Iteration $T$, step sizes $\eta_{0}$, $\eta_{1}$, and $\eta_{2}$, batch size $n$, meta batch size $m$, bound $\epsilon$, \#iterations $K$ in inner optimization, 
        classifier network $f_{_{\boldsymbol{W}}}$, weighting network $f_{\boldsymbol{\Omega}}$, 
       $D^{\text{train}}$, $D^{\text{meta}}$.\\
       \textbf{Output}: Trained robust network $f_{_{\boldsymbol{W}}}$.
    \begin{algorithmic}[1]

\STATE {Initialize networks $f_{_{\boldsymbol{W}}}$ and $f_{\boldsymbol{\Omega}}$;}
\FOR {$t = 1$ to $T$}
    \STATE{Sample $n$ and $m$ samples from $D^\text{train}$ and $D^\text{meta}$;}
    \FOR{$i=1$ to $n$ (in parallel)}
    \STATE{$\boldsymbol{x}_{i}^{\text{adv}}\!=\!\boldsymbol{x}_{i}\!+\!0.001\mathcal{N}(0,I)$ and $\boldsymbol{x}_{i}^{\text{at-adv}} \!=\!\boldsymbol{x}_{i}\!+\!0.001\mathcal{N}(0,I)$, where $\mathcal{N}(0,I)$ is the Gaussian distribution;}
    \STATE{Calculate the perturbation bound $\epsilon_{i}$ for sample $\boldsymbol{x}_{i}$;}
    \FOR {$k = 1$ to $K$}
    \STATE{$\boldsymbol{x}_{i}^{\text{adv}}\!\!\!\leftarrow\!\!\! \Pi_{\mathbb{B}(\boldsymbol{x}_{i}, \epsilon_{i})}(\eta_{0} \!\operatorname{sign}(\nabla_{\boldsymbol{x}_{i}^{\text{adv}}} \ell(f_{_{\boldsymbol{W}}}(\boldsymbol{x}_{i}), f_{_{\boldsymbol{W}}}(\boldsymbol{x}_{i}^{\text{adv}})))$\\$+\boldsymbol{x}_{i}^{\text{adv}})$, where $\Pi$ is the projection operator;}
    \STATE{$\boldsymbol{x}_{i}^{\text{at-adv}}\!\!\!\leftarrow\!\!\! \Pi_{\mathbb{B}(\boldsymbol{x}_{i}, \epsilon_{i})}(\!-\!\eta_{0} \!\operatorname{sign}(\nabla_{\boldsymbol{x}_{i}^{\text{at-adv}}} \ell(f_{_{\boldsymbol{W}}}(\boldsymbol{x}_{i}^{\text{at-adv}}), y_{i}))$\\$+\boldsymbol{x}_{i}^{\text{at-adv}})$;}
    \ENDFOR
    \ENDFOR
        \STATE{Formulate 
    $\hat{\boldsymbol{W}}^{(t)}(\boldsymbol{\Omega})$ by Eq.~(\ref{meta1});}
     \STATE{Update 
    $\boldsymbol{\Omega}^{(t+1)}$ by Eq.~(\ref{meta2}) and update 
    $\boldsymbol{W}^{(t+1)}$ by Eq.~(\ref{meta3});}
    \ENDFOR
    \end{algorithmic}
\end{algorithm}

\subsection{Extraction of Training Characteristics ($\boldsymbol{\zeta}_{\boldsymbol{x}}$)}
Our theoretical investigation reveals that different training samples can have different perturbation directions. The perturbation direction of a training sample depends on a series of factors, including learning difficulty, class proportion, and noise degree. Therefore, six training characteristics of each training sample $\boldsymbol{x}$, namely, loss ($\boldsymbol{\zeta}_{{\boldsymbol{x}},1}$), margin ($\boldsymbol{\zeta}_{{\boldsymbol{x}},2}$), the norm of loss gradient for the logit vector ($\boldsymbol{\zeta}_{{\boldsymbol{x}}, 3}$), 
the information entropy of the softmax output ($\boldsymbol{\zeta}_{{\boldsymbol{x}}, 4}$),
class proportion ($\boldsymbol{\zeta}_{{\boldsymbol{x}}, 5}$), and the average loss of each class ($\boldsymbol{\zeta}_{{\boldsymbol{x}},6}$), are extracted, as shown in the extraction module in Fig.~2. The calculation detail of each characteristic is shown in the online material.

\subsection{Perturbation Bound ($\epsilon_{\boldsymbol{x}}$) Calculation}

We employ two types of varied bound in our framework. Following Ref.~Xu et al.\shortcite{xu2021robust}, the class-wise perturbation bound named ReMargin, which is suitable for imbalanced data, is utilized.  A sample-wise bound is proposed to handle noise. It is inspired by the intuition that noisy samples have a large norm of loss gradient in general and these samples should exhibit the greatest degree of anti-adversarial training. Thus, the Grad-Based bound can be calculated as
\begin{equation}
    \epsilon_{\boldsymbol{x}} = (\alpha_{\boldsymbol{x}}\overline{\boldsymbol{g}}_{{\boldsymbol{x}}_{\text{adv}}}+\beta_{\boldsymbol{x}}\overline{\boldsymbol{g}}_{{\boldsymbol{x}}_{\text{at-adv}}}+\varepsilon)\times \epsilon,
\end{equation}
where $\overline{\boldsymbol{g}}_{{\boldsymbol{x}}_{\text{adv}}}$ and $\overline{\boldsymbol{g}}_{{\boldsymbol{x}}_{\text{at-adv}}}$ are the normalized $||\frac{\partial\ell(f_{_{\boldsymbol{W}}}(\boldsymbol{x}), f_{_{\boldsymbol{W}}}(\boldsymbol{x}_{\text{adv}}))}{\partial \boldsymbol{x}_{\text{adv}}}||_2$ and $||\frac{\partial \ell(f_{_{\boldsymbol{W}}}(\boldsymbol{x}_{\text{at-adv}}), y)}{\partial\boldsymbol{x}_{\text{at-adv}}}||_2$, respectively. 
$\epsilon$ is a predefined perturbation bound, and $\varepsilon$ is a hyperparameter that is set to $0.9$ in our experiments. This bound is also effective on imbalanced data because samples in tail classes have large norms of loss gradient, and they should do the greatest degree of adversarial training.

\subsection{Training with Meta-Learning}
On the basis of the extracted characteristics and calculated bounds, an online learning strategy 
is adopted to alternatively update $\boldsymbol{W}$ and $\boldsymbol{\Omega}$ using a single optimization loop, as shown in Fig.~2. Assume that we have a small amount of unbiased meta data $D^{\text{meta}}\!=\!\{\boldsymbol{x}^\text{meta}_{i},y^\text{meta}_{i}\}_{i=1}^{M}$, where $M \ll N$. Even if meta data are lacking, they can be compiled from the training data $D^{\text{train}}$~\cite{construct_meta}. The main steps are shown below. Here, we ignore the regularization terms introduced by the fairness constraints, while the online material provides the complete formulas. 

$\boldsymbol{\Omega}$ is treated as the to-be-updated parameter, and the parameter of the updated classifier $\boldsymbol{W}$, which is a function of $\boldsymbol{\Omega}$, is formulated. Stochastic gradient descent (SGD) is utilized to optimize the training loss.
Specifically, a minibatch of training samples $\{\boldsymbol{x}_{i},{y}_{i}\}_{i=1}^{n}$ is selected in each iteration, where $n$ is the size of the mini-batch. 
The updating of $\boldsymbol{W}$ can be formulated as 
\begin{equation}
\resizebox{0.9\hsize}{!}{$\begin{aligned}
&\hat{\boldsymbol{W}}^{(t)}(\boldsymbol{\Omega})=\boldsymbol{W}^{(t)}-\eta_{1} \frac{1}{n} \sum\nolimits_{i=1}^{n} \nabla_{_{\boldsymbol{W}}}\{\alpha_{i}[ \ell(f_{_{\boldsymbol{W}}}(\boldsymbol{x}_{i}), y_{i})+\\&\lambda \ell( f_{_{\boldsymbol{W}}}(\boldsymbol{x}_{i}),f_{_{\boldsymbol{W}}}(\boldsymbol{x}_{i}^{\text{adv}}))]+\beta_{i} \ell(f_{_{\boldsymbol{W}}}(\boldsymbol{x}_{i}^{\text{at-adv}}), y_{i})\}|_{_{\boldsymbol{W}^{(t)}}},
\end{aligned}$}
\label{meta1}
\end{equation}
where $\eta_{1}$ is the step size. 
The parameter of the weighting network
$\boldsymbol{\Omega}$ after receiving feedback from the classifier network can be updated on a minibatch of meta data as follows:
\begin{equation}
\resizebox{1\hsize}{!}{$\begin{aligned}
\boldsymbol{\Omega}&^{(t+1)}= \boldsymbol{\Omega}^{(t)}-\eta_{2} \frac{1}{m} \sum\nolimits_{i=1}^{m} \nabla_{\boldsymbol{\Omega}}[\ell^{\text {meta}}(f_{_{\hat{\boldsymbol{W}}^{(t)}(\boldsymbol{\Omega})}}(\boldsymbol{x}_{i}), y_{i})\\&+\lambda \ell^{\text {meta}}(f_{_{\hat{\boldsymbol{W}}^{(t)}(\boldsymbol{\Omega})}}(\boldsymbol{x}_{i}),f_{_{\hat{\boldsymbol{W}}^{(t)}(\boldsymbol{\Omega})}}(\boldsymbol{x}_{i}^{\text{adv}}))+\ell^{\text {meta}}(f_{_{\hat{\boldsymbol{W}}^{(t)}(\boldsymbol{\Omega})}}(\boldsymbol{x}_{i}^{\text{at-adv}}), y_{i})]|_{\boldsymbol{\Omega}^{(t)}},
\end{aligned}$}
\label{meta2}
\vspace{0in}
\end{equation}
where $m$ and $\eta_{2}$ are the minibatch size of meta data and the step size, respectively. The parameters of the classifier network are updated with the obtained weights by fixing the parameters of the
weighting network as $\boldsymbol{\Omega}^{(t+1)}$:
\begin{equation}
\resizebox{0.85\hsize}{!}{$\begin{aligned}
&\boldsymbol{W}^{(t+1)}=\boldsymbol{W}^{(t)}-\eta_{1} \frac{1}{n} \sum\nolimits_{i=1}^{n}\nabla_{_{\boldsymbol{W}}}\{\alpha_{i}[ \ell(f_{_{\boldsymbol{W}}}(\boldsymbol{x}_{i}), y_{i})+\\&\lambda \ell( f_{_{\boldsymbol{W}}}(\boldsymbol{x}_{i}),f_{_{\boldsymbol{W}}}(\boldsymbol{x}_{i}^{\text{adv}}))]+\beta_{i} \ell(f_{_{\boldsymbol{W}}}(\boldsymbol{x}_{i}^{\text{at-adv}}), y_{i})\}|_{_{\boldsymbol{W}^{(t)}}}.
\end{aligned}$}
\label{meta3}
\end{equation}
The steps of our CAAT method are shown in Algorithm~1.
\section{Experiments}
\begin{table*}[t]
\footnotesize
\centering
\begin{tabular}{l|cc|cc|cc}
\toprule
                               & Avg. Nat.      & Worst Nat.    & Avg. Bdy.    & Worst Bdy.    & Avg. Rob.     & Worst Rob.    \\ \midrule\midrule
PGD Adv. Training                          & 15.5          & 33.8          & 40.9          & 55.9          & 56.4          & 82.7          \\
TRADES ($1/\lambda=1$)                       & \underline{14.6}          & 31.2          & 43.1          & 64.6          & 57.7          & 84.7          \\
TRADES ($1/\lambda=6$)                        & 19.6          & 39.1          & 29.9          & 49.5          & 49.3          & 77.6          \\
Baseline ReWeight & 19.2 & 28.3 & 39.2 & 53.7 & 58.2 & 80.1\\
FRL (ReWeight)            & 16.0          & \textbf{22.5} & 41.6          & 54.2          & 57.6          & 73.3          \\
FRL (ReMargin)            & 16.9          & 24.9          & 35.0          & 50.6          & 51.9          & 75.5          \\
FRL (ReWeight+ReMargin) & 17.0          & 26.8          & 35.7          & 44.5          & 52.7          & 69.5          \\ \hline
CAAT (Grad-Based)      & \underline{14.6} &	\underline{23.6} &	\textbf{14.4} &	\textbf{23.3} &	\textbf{28.6} &	\underline{48.1}
    \\
CAAT (ReMargin)      & \textbf{13.9} &	24.3	& \underline{15.4}	& \underline{24.9} &	\underline{29.3} &	\textbf{44.4}

\\ \bottomrule
\end{tabular}
\caption{Average and worstclass natural, boundary, and robust errors (\%) for various algorithms on CIFAR10.}
\end{table*}

\begin{figure}[b] 
\centering
\includegraphics[width=0.46\textwidth]{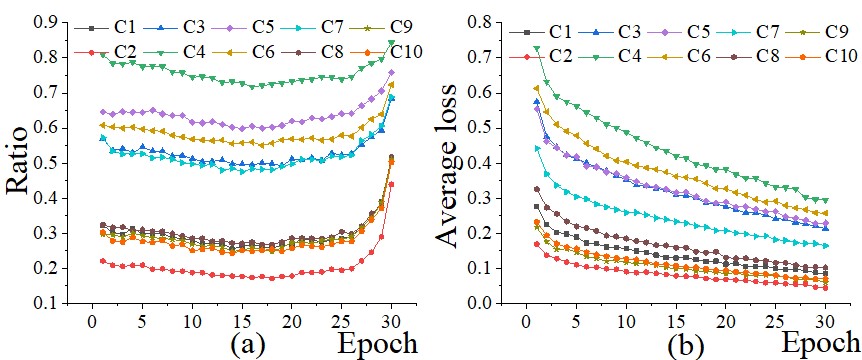}
\caption{(a): Ratio of adversaries in each class during training on standard CIFAR10. (b): Average loss of each class during training on standard CIFAR10.}
\end{figure}
\begin{table}[bp]
\footnotesize
\centering
\begin{tabular}{l|c|c|c}
\toprule
                               & Avg. Nat.      & Avg. Bdy.   & Avg. Rob.  \\ \midrule\midrule
PGD Adv. Training                           & 15.6 & 37.1 & 52.8

         \\
TRADES ($1/\lambda=1$) &                       15.6& 31.0&	46.5

         \\
TRADES ($1/\lambda=6$)                        & 16.4& 21.0& 37.4

         \\
FRL (ReWeight)            & 15.3 &	36.0&	51.4 

   \\
FRL (ReMargin)        &  15.2 &	36.0&	51.1 

         \\
\scriptsize{FRL (ReWeight+ReMargin)} & 15.7 &	34.3	&	50.0

        \\ \hline
CAAT (Grad-Based)     & \textbf{14.6} &	\textbf{13.9}&	\textbf{28.5}
    \\
CAAT (ReMargin)   & \underline{14.7}	& \underline{14.7} &	\underline{29.4}

\\ \bottomrule
\end{tabular}
\caption{Average natural, boundary, and robust errors (\%) for various algorithms on CIFAR10 with 20\% pair-flip noise.}
\end{table}
Experiments are conducted to verify our theoretical findings and the effectiveness of the proposed CAAT in
improving the accuracy, robustness, and fairness of the robust models. 
\subsection{Experimental Settings}
Benchmark adversarial learning datasets: CIFAR10~\cite{cifar10} and SVHN~\cite{svhn} are adopted in our experiments, including the noisy and imbalanced versions of the CIFAR data~\cite{shu2019meta}. 
For the two datasets, PreAct-ResNet18~\cite{pre-resnet} and Wide-ResNet28-10 (WRN28-10)~\cite{wide-resnet} are adopted as the backbone network. This section only represents the results of PreAct-ResNet18. Others are presented in the online material. 
The compared methods include three popular adversarial
training algorithms, namely, PGD~\cite{pgd}, TRADES~\cite{zhangtrades}, and FRL~\cite{xu2021robust}. A debiasing method~\cite{compared_method} is also compared which is to upweight the loss of the class with the largest robust error in the training data. The results of TRADES and FRL are calculated by using the codes in their official repositories. 

The training and testing configurations used in Ref.~Xu et al. \shortcite{xu2021robust} are followed. The number of iterations in an adversarial attack is set to 10. Following Xu et al.~\shortcite{xu2021robust}, 300 samples in each class with clean labels are selected as the meta dataset, which helps us tune the hyperparameters and train the weighting network. Adversarial
training is trained on PGD attack setting $\epsilon = 8/255$ with cross-entropy loss. For our method and FRL (ReMargin), the predefined perturbation bound is also set to $8/255$. All the models are trained by using SGD with
momentum 0.9 and weight decay $5\times10^{-4}$. 
The value of $\lambda$ is selected in $\{2/3,1,1.5,6\}$. 
During the evaluation
phase, we report each model's average and worstclass natural, boundary, and robust error rates.

\begin{figure}[b] 
\centering
\includegraphics[width=0.46\textwidth]{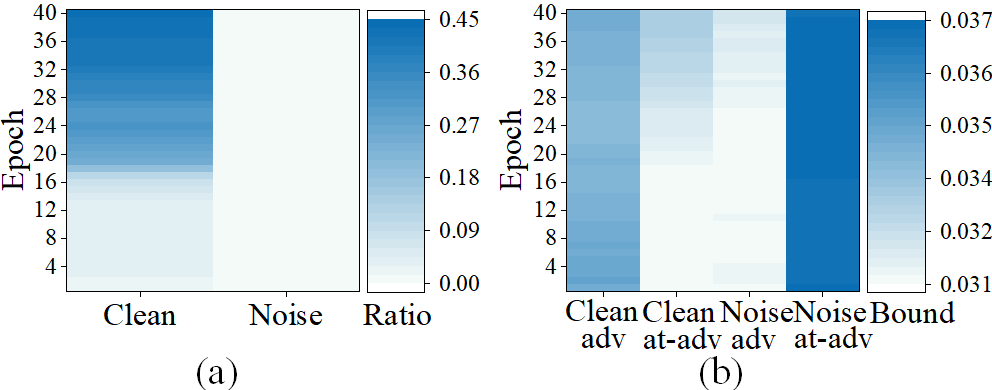}
\caption{(a): Ratio of adversaries for noisy and clean samples on CIFAR10 with 20\% uniform noise during training. (b): Average adversarial and anti-adversarial perturbation bounds for clean and noisy samples during training.}
\end{figure}

\subsection{Experiments on Standard Dataset}
Tables~1 shows the performance of our proposed CAAT and the compared methods on standard CIFAR10. Those on SVHN are shown in the online material. Considering that our training/testing configuration is the same as that in Ref.~Xu et al.\shortcite{xu2021robust}, the results of the above competing
methods reported in the FRL~\cite{xu2021robust} paper are
directly presented.

\begin{figure*}[t] 
\centering
\includegraphics[width=0.92\textwidth]{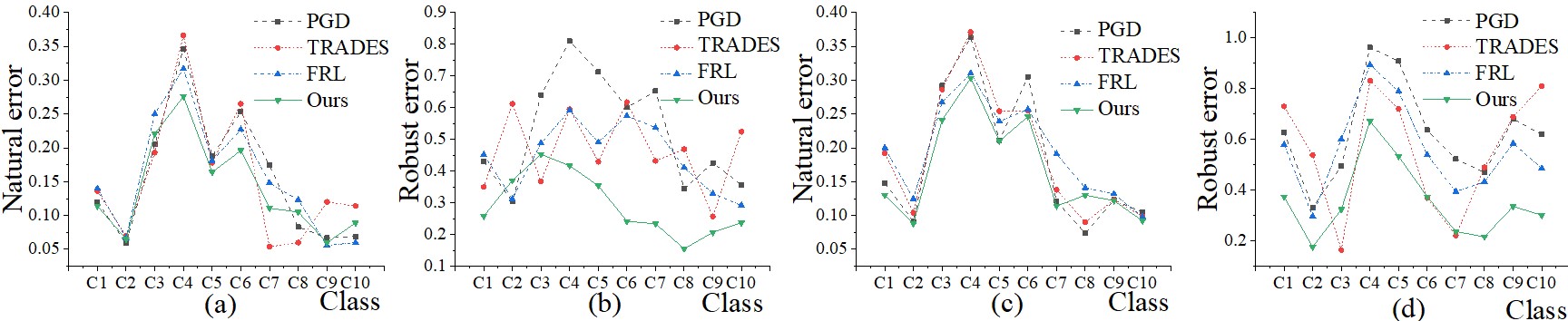}
\caption{(a) and (b): Natural and robust errors for each class of different methods on CIFAR10 with imbalance factor 10. (c) and (d): Natural and robust errors for each class of different methods on CIFAR10 with imbalance factor 100.}
\end{figure*}


From the results, our methods with two types of bound reduce the average natural and robust
errors under different degrees, indicating that CAAT obtains better accuracy and robustness of the model.
Compared with other methods, 
CAAT decreases the average and worst robust error rates by 21\% and 25\% on CIFAR10. Baseline ReWeight can only decrease the worst intraclass natural error but cannot equalize boundary or robust errors. FRL~\cite{xu2021robust} has only a limited ability to reduce the worst boundary and robust errors, resulting in limited fairness between classes. 
Our method more effectively decreases the worst intraclass errors.
Thus, CAAT achieves better fairness among classes compared with other methods. Although FRL (ReWeight) obtains the lowest worst natural error, it has large average and worst robust errors, which is inferior to CAAT.
Hard classes (classes with 
a large average loss) have a higher ratio of adversaries than easy ones, as shown in Fig.~3, which helps improve the performance of hard classes and effectively enhances the fairness among classes. The same conclusions can also be obtained on the SVHN dataset. 

\begin{table}[t]
\footnotesize
\centering
\begin{tabular}{l|c|c|c}
\toprule
                               & Avg. Nat.        & Avg. Bdy.    & Avg. Rob.     \\ \midrule\midrule
PGD Adv. Training                           & 20.1 &		42.8 &	62.9 
         \\
TRADES ($1/\lambda=1$) &                       16.8 &	32.3 &	49.1	
         \\
TRADES ($1/\lambda=6$)                        & 23.6 &		23.8 &	47.4 
         \\
FRL (ReWeight)            & 16.9 &	38.1 &		55.0 
   \\
FRL (ReMargin)           & 17.5		& 35.6 &	53.1
         \\
\scriptsize{FRL (ReWeight+ReMargin)} & 17.2  &	35.1 &		52.3 
        \\ \hline
CAAT (Grad-Based)      & \textbf{15.8} &	\underline{14.2} &		\underline{30.0}

    \\
CAAT (ReMargin)    &  \underline{16.2}	 &  	\textbf{13.7}  &	\textbf{29.9}

\\\bottomrule
\end{tabular}
\caption{Average and worstclass natural, boundary, and robust errors (\%) on CIFAR10 with imbalance factor 10.}
\end{table}

\subsection{Experiments of Noisy Classification}
Two settings of corrupted labels, including uniform and pair-flip noises, are adopted~\cite{shu2019meta}. 
The values of the noise ratio are set to $20\%$ and $40\%$. CIFAR10 dataset, which is popularly used for the evaluation of noisy labels, is adopted. Here, we only show the average errors of CIFAR10 with $20\%$ pair-flip noise. Others are presented in the online material. From the results in Table~2 and the online material, CAAT achieves the lowest average and worst natural and robust errors,
indicating that it obtains the best generalization, robustness, and fairness compared with other methods.

As shown in Fig.~4~(a), most of the noisy samples are anti-adversarially perturbed during training, which is in accordance with our theoretical findings. From Fig.~4~(b), the average anti-adversarial perturbation bound for noisy samples is the largest, implying that noisy samples exhibit the largest degree of anti-adversarial training. Thus, the negative influence of noisy samples can be decreased.
The ratio of adversaries for clean samples increases with the progress of training,
demonstrating that clean samples are playing a more important role than noisy ones during training.

\subsection{Experiments of Imbalanced Classification}

The long-tailed version of CIFAR10 compiled by Cui et al.~\shortcite{imbalanced_data} is utilized. The values of the imbalance factor are set to $10$ and $100$. Here, we only show the average results when the imbalance factor equals $10$. Others are presented in the online material. Compared with other methods, CAAT achieves the minimum average and worst natural and robust errors, as shown in Table~3. 
As shown in Fig.~5, CAAT decreases the natural and robust errors for most classes and achieves the lowest performance gap among different classes. We also verify that the first head class has the lowest ratio of adversaries and tail classes have a high ratio of adversaries, which is consistent with our theoretical findings. 
The details are presented in the online material.  

\begin{table}[t]
\footnotesize
\centering
\begin{tabular}{l|c|c|c}
\toprule
                                    & Avg. Nat. (\%) & Avg. Bdy. (\%) & Avg. Rob. (\%)    \\ \midrule\midrule
Setting I          & 16.0      & 41.6         & 57.6       \\
Setting II      & 16.1    & 35.8      & 51.9         \\
Setting III      & \underline{14.9}    & \textbf{13.8}  & \textbf{28.7}  \\
Setting IV & \textbf{13.9}    &  \underline{15.4}    &  \underline{29.3}      \\ \midrule
\end{tabular}
\caption{Ablation studies of CAAT on standard CIFAR10.}
\end{table}
\subsection{Ablation Studies}
Four variations of CAAT are considered, including adversarial training with the same perturbation direction and bound (Setting~I), adversarial training with the same perturbation direction and different bounds (Setting~II), adversarial training with different perturbation directions (adversaries and anti-adversaries) and the same bound (Setting~III), and adversarial training with different perturbation directions and bounds (Setting~IV). PreAct-ResNet18 is used. The results are shown in Table~4.
Settings~III and~IV obtain better performance compared with Settings~I and~II. Thus, the combination strategy is more effective. Compared with Setting~III, Setting~IV further decreases the average natural error, indicating that the varied bound is more valid in some cases. The worst errors are shown in the online material.

\section{Conclusions}
This study theoretically investigates the role of adversarial training with different directions (adversarial and anti-adversarial) and bounds for the robust model. 
Three typical occasions are considered, including 
classes with different difficulties, imbalance learning, and noisy label learning. A series of theoretical findings are obtained, illuminating a new objective function that combines adversaries and anti-adversaries in training. Consequently, an adversarial training framework (CAAT) is proposed to solve the objective, in which meta learning is utilized to optimize the combined weights of the adversary and anti-adversary for each sample in accordance with its learning characteristics. 
Extensive experiments verify the rationality of our theoretical findings and the effectiveness of CAAT in achieving better accuracy, robustness, and fairness of the robust models compared with other adversarial training methods. 

\section{Acknowledgments}
This study is partially supported by NSFC 62076178, TJF 22ZYYYJC00020, and 19ZXAZNGX00050.

\bibliography{aaai23}

\end{document}